# Organization of Valence and Arousal in Vision–Language Representations of Built Environments: Insights from the EMOIS Dataset

Madoka Yonekura, Katsunori Kohda, Nobuhiko Muramoto and Takahiro Yamaguchi

***Abstract*—Visual perception of built environments contributes to the affective impressions that people form in everyday life. However, how these impressions are represented within vision foundation models remains largely unexplored. To support the systematic investigation of this subject, we introduce the Emotional Impression of Spaces (EMOIS) dataset, comprising 1,544 real-world built-environment images. Each image is annotated with image-evoked valence and arousal ratings collected from Japanese adults by conducting a large-scale web-based survey, with approximately 120 ratings per image. Using Contrastive Language–Image Pre-training (CLIP) representations, we perform predictive and geometric analyses to systematically investigate how valence and arousal are encoded and organized within the representation space. These analyses reveal that valence exhibited stronger and more coherent organization than arousal. Cross-dataset analyses with the Open Affective Standardized Image Set (OASIS), a benchmark dataset of general affective photographs, reveal differences in affective organization between the two datasets. Regression analyses demonstrate high predictive performance for valence and arousal within EMOIS, with mean coefficients of determination of 0.865 and 0.807, respectively, across repeated internal hold-out evaluations. Finally, we present an example-based interface illustrating how learned representations can support qualitative interpretation of predicted affective values. These findings can help elucidate affective representations of built environments and establish EMOIS as a densely annotated resource for future affective computing research in this domain.**

## I. Introduction

BUILT environments evoke emotional responses via visual perception and can influence emotion, cognition, and psychological well-being [1]–[4]. Because people spend a substantial portion of their lives in built environments like homes, workplaces, and public spaces [5], understanding and modeling the affective impressions of built environments is important for affective computing, human-centered design, and environmental evaluation [2], [6].

Research on affective computing has demonstrated that emotional responses can be modeled from visual stimuli using data-driven approaches [7]–[11]. More recently, vision–language foundation models have been successfully used to model subjective perceptual and affective judgments in response to images [12]–[15]. These studies have suggested that learned representations encode information that is relevant to human affective perception. However, they have primarily focused on predictive performance, without sufficiently clarifying how affective information is organized within learned representations.

Moving beyond predictive performance can enhance our understanding of affective representations in foundation models, thereby advancing affective computing. Thus, an open question arises: how is affective information organized within foundation-model representations? Specifically, it remains unclear whether different affective dimensions exhibit similar or distinct representational organizations, and whether these organizations are preserved across different visual domains. Answering this question requires affective datasets with sufficient coverage of the target visual domain.

Existing affective image datasets have substantially advanced visual affect analysis; however, dataset diversity and generalization across visual domains remain limited [16]. Most existing datasets primarily consist of general scenes, objects, or events, providing relatively limited coverage of built environments [17]–[19]. Consequently, they only provide limited opportunities to investigate representational organization in built environments. A dedicated affective dataset covering diverse built environments is needed to support systematic investigation of affective representations in this domain.

In this study, we introduce Emotional Impression of Spaces (EMOIS), a new affective dataset of various built environments. Using Contrastive Language–Image Pre-training (CLIP) [20], we first investigate whether affective impressions of built environments can be reliably predicted within EMOIS from vision–language representations. We then analyze how affective information is organized within the resulting representation space.

The main contributions of this study are summarized as follows:

1) **Construction of EMOIS:** We introduce EMOIS, a dataset containing 1,544 real-world built-environment images annotated with valence and arousal ratings, which provides a benchmark for studying affective responses to built environments.
2) **CLIP-based affective prediction for built environments:** We demonstrate that affective impressions of built environments can be predicted within EMOIS

The authors are with Toyota Central R&D Labs., Inc., Nagakute, Aichi 480-1192, Japan (e-mail: t-yamaguchi@mosk.tytlabs.co.jp).

using CLIP representations and a lightweight regression framework.

3) **Analysis of affective organization across dimensions and domains:** We analyze how valence and arousal are organized within CLIP representations of built environments and compare their representational organization with those of general affective image datasets.
4) **Example-based interpretation of affective predictions:** We present an example-based retrieval interface that supports interpretation of affective predictions by presenting visually similar reference spaces together with their subjective affective ratings.

## II. Related Work

### A. Emotion Representation Models

Emotions are commonly represented using categorical or dimensional models. Categorical models describe emotions as discrete classes like happiness, sadness, anger, and fear [21], [22], while dimensional models represent emotions within a continuous affective space. Early dimensional approaches included the pleasure–arousal–dominance (PAD) model [23]. Subsequently, Russell introduced the valence–arousal (VA) circumplex model, in which valence represents the degree of pleasantness or unpleasantness and arousal reflects the level of physiological or psychological activation [24]. The VA framework has been widely adopted in psychology, affective computing, and environmental research to model emotional responses to visual stimuli and spatial environments [4], [25], [26].

Recent studies have suggested that categorical and dimensional models are not necessarily mutually exclusive. For example, Cowen and Keltner demonstrated that emotional experiences can be organized into numerous emotion categories connected via continuous affective gradients [27]. This finding indicates that emotional experiences may be structured in ways that are not fully captured by either discrete emotion categories or low-dimensional affective representations alone.

### B. Affective Computing and Representation Learning

Affective computing based on visual data has evolved from handcrafted features to modern representation-learning approaches [7]–[9]. Early studies relied on low-level visual descriptors like color, texture, and composition [7], whereas later approaches incorporated higher-level semantic representations and deep learning models to infer emotional information directly from large-scale image collections [8], [9].

Recent advances in vision–language models, particularly CLIP [20], have further expanded representation-learning approaches for visual understanding. CLIP, which is pre-trained via contrastive learning on large-scale image–text datasets, provides general-purpose vision–language representations that have proven useful across a wide range of downstream tasks. Hentschel et al. demonstrated that simple linear models trained on CLIP embeddings achieve competitive performance in image aesthetic assessment [12]. Similarly, Wang et al. showed that CLIP representations capture subjective perceptual attributes related to the "look and feel" of images, suggesting that high-level perceptual characteristics are reflected within CLIP embedding spaces [13]. Notably, Conwell et al. reported that a wide range of visual models can be used to explain a substantial proportion of visually evoked human affective responses, suggesting that perceptual representations may contain information relevant to affective judgments even without explicit affective supervision [15].

Collectively, these studies have suggested that modern foundation representations contain information relevant to subjective human evaluation and affective judgments, motivating further investigation into how such information is organized within representation spaces. However, the existing evidence has primarily been obtained from general photographic imagery and object-centric datasets. Whether similar relationships hold for architectural and built environments remains largely unexplored.

Affective image datasets based on the VA framework have played a central role in the study of emotional responses to visual stimuli. Representative datasets include the International Affective Picture System (IAPS) [18], Nencki Affective Picture System (NAPS) [19], and Open Affective Standardized Image Set (OASIS) [17], all of which provide image valence and arousal annotations. These datasets have been extensively used in psychology and affective computing research; however, they primarily consist of object-centric or general photographic imagery and do not adequately capture the structural, compositional, and environmental characteristics of built environments.

Moreover, although large-scale datasets exist for scene understanding and subjective environmental evaluation, including Place Pulse [28], ScenicOrNot [29], Places [30], and SUN RGB-D [31], they do not provide affective annotations based on established emotional dimensions such as valence and arousal.

Consequently, it remains unclear whether modern foundation representations encode affective characteristics specific to built environments and how such affective information is organized within representation geometry.

### C. Emotional Perception of Architectural Space

Emotional perception in architectural environments has been investigated in several studies [1]–[3], [32], [33]. These studies have shown that both geometric and perceptual characteristics of built environments influence affective responses. Geometric properties such as curvature, enclosure, ceiling height, and spatial proportions are associated with differences in affective evaluation [25], [34], [35]. For example, curved forms are generally perceived as more pleasant than rectilinear designs, whereas higher ceilings have been linked to greater feelings of freedom and a positive affect [32], [34], [35]. Variations in color properties, lighting conditions, and material appearance have been shown to influence perceived pleasantness and arousal [36]–[38]. These findings suggest that emotional impressions emerge from the combined influence of multiple architectural attributes rather than from isolated design elements.

To investigate such relationships under controlled conditions, many studies have employed virtual reality environments that allow individual spatial variables to be manipulated while minimizing confounding factors [25], [39], [40]. Such studies have provided valuable evidence regarding the causal effects of specific design attributes on emotional responses. Recently, Xylakis et al. [26] extended this line of research by investigating emotional responses during navigation via virtual room sequences, demonstrating that an affective experience evolves over time as people move through different spaces. However, these approaches typically rely on controlled or simplified virtual environments.

Gregorians et al. [41] used first-person-view videos of real architectural spaces and demonstrated that experiential dimensions like fascination, coherence, and hominess are closely related to the affective evaluations of architectural environments. Collectively, these findings suggest that affective experience in architecture emerges from individual design attributes as well as the sequence of spaces encountered during the architectural experience.

More recently, Aseniero et al. [42] fine-tuned CLIP using photographs and 3D-rendered scenes from an office environment to support human-centered evaluation of designed spaces. They evaluated the scenes in terms of three experiential dimensions—social, tranquil, and inspirational—and integrated the resulting predictions into interactive floorplan and 3D visualization tools. Despite these advances, the computational modeling of emotional impressions from large collections of diverse real-world built-environment images remains relatively underexplored.

### D. Example-Based Interpretation of Affective Predictions

Various explainable AI approaches have been proposed to improve the transparency of predictive models. A recent review of explainable AI in affective computing categorized these approaches into feature-, concept-, example-, rule-, and modality-based methods [43].

Feature-based approaches include SHapley Additive exPlanations (SHAP), which estimates feature contributions to model outputs [44], and Gradient-weighted Class Activation Mapping (Grad-CAM), which visualizes image regions that influence predictions [45]. In addition to identifying influential features or regions, interpretability can be further enhanced by enabling the interpretation of the features themselves [43].

More recently, emotion-focused frameworks have also been developed. For example, the Emotion Stimuli Segmentation and Explanation Model (EmoSEM) identifies emotion-evoking visual regions and generates textual explanations of emotional responses [46].

Example-based approaches provide another form of interpretability by presenting representative, similar, or counterfactual examples that offer intuitive information about model decisions [43]. For example, prototype-based methods such as ProtoPNet explain predictions through comparisons with representative training examples [47]. In affective prediction, providing examples along with subjective affective ratings can provide concrete reference points for interpreting predicted values.

Building on this perspective, we propose an example-based interpretation framework that complements conventional explainability methods by enabling affective predictions to be interpreted through comparisons with visually similar reference images and their associated subjective affective ratings.

## III. EMOIS Dataset

EMOIS is an affective dataset of built environments images, consisting of 1,544 images representing a wide variety of indoor and outdoor spaces encountered in everyday life. Emotional impressions were annotated in terms of valence and arousal based on the VA dimensional framework [24], and annotations were collected using a seven-point Likert scale. Details of the image collection, preprocessing, and annotation are described in the following subsections.

Because the dataset includes environments characteristic of Japanese architecture, such as traditional Japanese-style rooms (*washitsu*) and shrines, participants were restricted to individuals whose nationality and longest residence were both in Japan. This restriction was introduced because affective evaluations in terms of valence and arousal may depend on cultural background and contextual experience, particularly for culturally specific architectural environments [48]–[50].

### A. Image Collection

Images were collected using a diverse set of keywords describing the functions of spaces (e.g., offices, living rooms, Japanese-style rooms (*washitsu*), schools, restaurants, vehicle interiors, cafés, and industrial workspaces) and atmospheric characteristics (e.g., derelict, empty, colorful, green-view, and illuminated spaces), based on commonly used terminology in architecture and interior design. This strategy was intended to ensure diversity in both functional purposes and affective atmospheres.

To further increase visual diversity, multiple interior design styles, including modern, Scandinavian, and natural, were included. Images containing vegetation like potted plants and floral decorations were also incorporated because these elements are common in built environments and may influence the affective impressions of spaces [51], [52]. Images containing people or textual content were excluded to ensure that emotional impressions were primarily derived from spatial characteristics rather than social or semantic cues.

To reduce redundancy, images that were recognized as depicting the same built environment were avoided during image collection, regardless of differences in viewpoint, composition, or photographed area. As an additional post-collection quality-control step, cosine similarity between CLIP image embeddings was computed, and image pairs with cosine similarity greater than 0.9 were visually inspected. This additional screening identified 70 images forming 33 groups depicting the same built environments. These images were retained in the final dataset but assigned common group identifiers for subsequent group-aware analyses. All images were obtained from the PIXTA image service under a commercial machine-learning license to ensure their suitability for commercial machine-learning applications.

### B. Image Preprocessing

Because CLIP [20] uses square image inputs, all images were converted into square format before annotation and feature extraction. Although CLIP internally resizes images to $224 \times 224$ pixels, directly presenting such low-resolution images in an annotation survey could reduce the perceptual quality and affect emotional evaluation. Therefore, all images were cropped into a square format using Adobe Photoshop (Adobe Inc.) and resized to $1{,}000 \times 1{,}000$ pixels for the web-based annotation survey. Cropping was performed individually for each image. The same square-cropped images were subsequently used for feature extraction to ensure consistency between the annotated stimuli and images used for affective prediction. Because of licensing restrictions, neither the original nor the square-cropped images can be publicly redistributed.

### C. Emotion Annotation

The emotional impressions of spatial images were evaluated using a VA dimensional model for emotion [24]. Participants rated either valence or arousal on a seven-point Likert scale. The annotation survey was conducted using a web-based questionnaire administered by ASMARQ Co., Ltd. Participants were recruited from the ASMARQ web panel and first completed a screening questionnaire that collected information on age, sex, nationality, and country of longest residence. Adults aged 20–69 years whose nationality and longest residence were both in Japan were eligible to proceed to the main survey. Explanations of the valence and arousal dimensions were presented in Japanese, with wording developed with reference to the annotation instructions used in the OASIS dataset [17]. To avoid confusion between the two emotional dimensions, each participant participated in only one survey session and evaluated only one dimension, either valence or arousal. To reduce response fatigue, each participant rated 50 images in random order.

Among the 50 images, six were fixed reference images used as gold-standard images. The remaining images were randomly divided into subsets of 44 images, and each participant was assigned one of these subsets. The presentation order of all 50 images was randomized for each participant. For each affective dimension, the same six gold-standard images were included in every survey session for all participants. The gold-standard images consisted of three high-scoring and three low-scoring images for the evaluated dimension (either valence or arousal), which were selected through a preliminary survey.

Responses were excluded if the main survey was completed in 1 min or less or 100 min or more, or if the entire survey, including screening and the main survey, was completed in 2 min or less or 100 min or more. These completion-time criteria were established before the response data were delivered to the authors. Responses in which the same rating was assigned to at least 48 of the 50 images within a session were also excluded. After applying the quality-control criteria, responses from 4,205 participants were retained for each affective dimension, with the final sample balanced across sex and age groups (420–421 participants in each sex-by-age-group stratum; 20s–60s). The study protocol was reviewed and approved by the Institutional Ethics Committee of Toyota Central R&D Labs., and informed consent was obtained from all participants prior to participation.

To assess the consistency of the collected ratings, a split-half reliability analysis was conducted separately for valence and arousal. The participants were randomly divided into two groups, and the mean image ratings were computed for each group. The Pearson correlation coefficients between the two sets of image ratings were calculated, and this procedure was repeated 1,000 times using different random splits. The mean correlation coefficient across the repetitions was used as the split-half reliability estimate.

## IV. Affective Representation Learning and Analysis Framework

Figure 1 provides an overview of the proposed framework. Spatial images are first encoded using a frozen CLIP model. The resulting representation space is subsequently used for affective prediction, representation analysis, and example-based interpretation of model outputs.

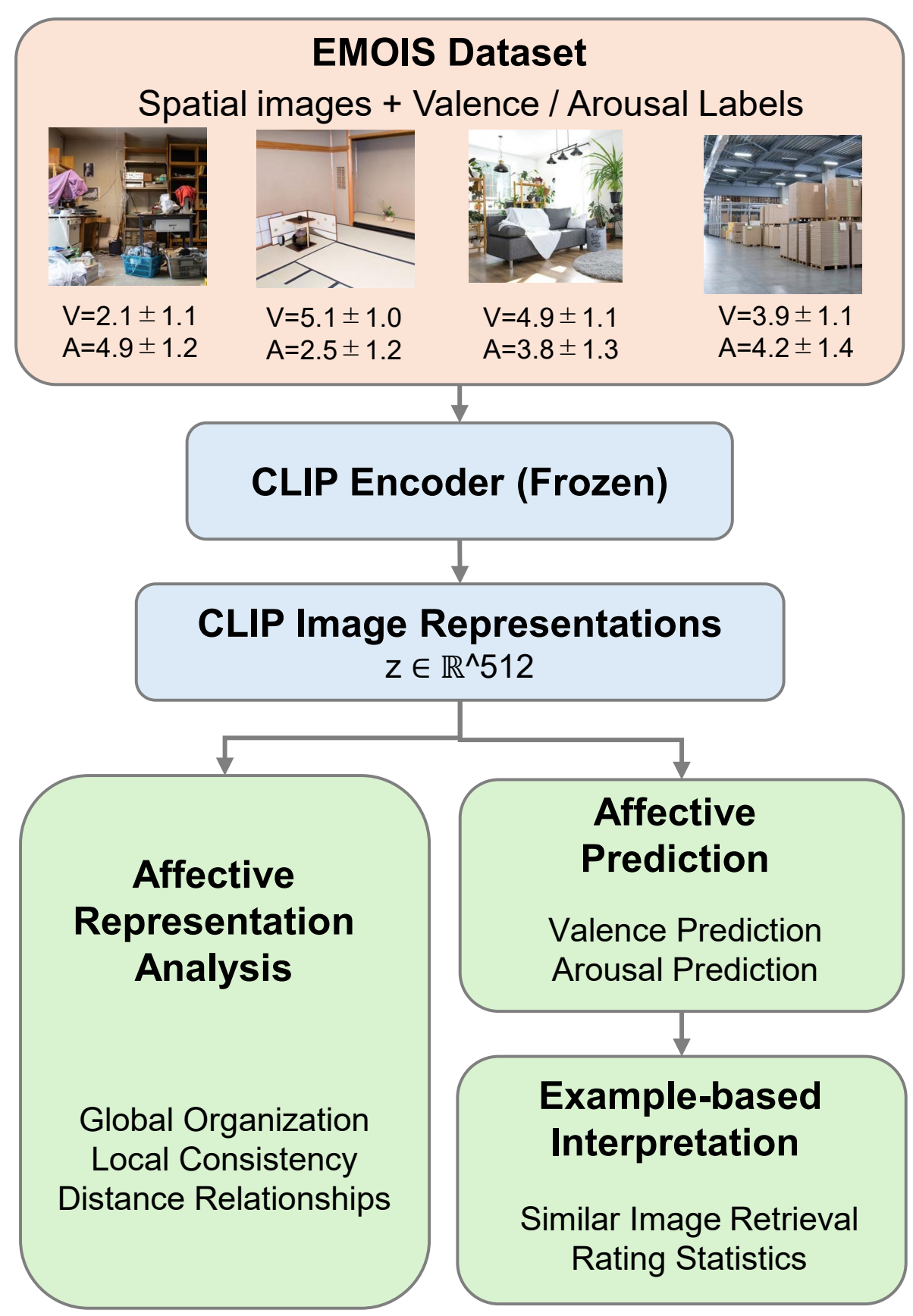

Fig. 1. Overview of the proposed framework. Spatial images from EMOIS annotated with valence and arousal ratings are encoded using a frozen CLIP model. The resulting image representations are used for affective prediction, representation analysis, and example-based interpretation through retrieval of similar reference images together with their subjective ratings.

### A. Spatial Emotion Prediction Model

Each image in the EMOIS dataset was converted into a feature representation using the CLIP image encoder (ViT-B/16) [20]. We used the publicly available OpenAI implementation of CLIP [53], and the same CLIP backbone (ViT-B/16) was employed in all experiments.

$$\mathbf{f}_i = E_{\text{CLIP}}(I_i), \tag{1}$$

where $I_i$ denotes the input image and $\mathbf{f}_i \in \mathbb{R}^{512}$ represents the extracted feature vector generated by the CLIP encoder. Before regression training, the extracted feature vectors were standardized using z-score normalization (zero mean and unit variance) based on the statistics computed from the training set. The same transformation parameters were subsequently applied to the test set.

For comparison, feature representations extracted using a standard supervised Vision Transformer (ViT; ViT-B/16) [54] and ResNet-based encoder [55] were also evaluated. Although the feature dimensionalities differed among models, dimensionality alignment and reduction were not applied. This choice was made to preserve the intrinsic information encoded in each pre-trained representation and to avoid altering its native geometry through dimensionality alignment or feature transformation.

The 1,544 images of the EMOIS dataset were divided into training and test sets using a ratio of 8:2. For image groups identified as depicting the same built environment (Section III-A), group-wise train–test splits were adopted such that all images within the same group were assigned entirely to either the training or the test set. A total of 100 group-wise train–test splits were generated using different random seeds and shared across all visual representations to ensure a fair comparison of prediction performance.

Bayesian ridge regression [56] was adopted to predict the mean valence and arousal ratings from the extracted feature vectors. This approach performs regularized linear prediction while preserving the original feature dimensionality and also provides predictive uncertainty estimates that can be incorporated into the proposed interpretation framework. Bayesian ridge regression was implemented using scikit-learn (version 1.7.2) with the default hyperparameter settings.

Two independent Bayesian ridge regression models were trained to predict valence and arousal:

$$\hat{v}_i = \mathbf{w}_v^\top \mathbf{f}_i + b_v, \tag{2}$$

$$\hat{a}_i = \mathbf{w}_a^\top \mathbf{f}_i + b_a, \tag{3}$$

where $\hat{v}_i$ and $\hat{a}_i$ denote the predicted mean valence and arousal scores, respectively. Moreover, $\mathbf{w}_v, \mathbf{w}_a \in \mathbb{R}^{512}$ are regression coefficient vectors, and $b_v$ and $b_a$ are bias terms.

The model performance was evaluated over the 100 pre-generated train–test splits. For each visual representation, the mean, maximum, and 2.5th–97.5th percentile interval of the coefficient of determination ($R^2$) for the test sets were calculated.

### B. Local Affective Consistency

To investigate the local neighborhood organization within each embedding space, retrieval analyses were conducted using ResNet50, ViT-B/16, and CLIP ViT-B/16 image representations. All EMOIS images were projected into their corresponding representation spaces and the cosine similarity between a query image and all remaining images was computed as follows:

$$\text{sim}(\mathbf{f}_i, \mathbf{f}_j) = \frac{\mathbf{f}_i^\top \mathbf{f}_j}{\|\mathbf{f}_i\| \, \|\mathbf{f}_j\|}. \tag{4}$$

For each query image, images belonging to the same built-environment group as the query image were excluded from the retrieval candidates. The five most similar images among the remaining candidates were then retrieved based on cosine similarity. The mean valence and arousal ratings of the retrieved images were averaged to obtain a retrieval-based affective estimate. Pearson correlation coefficients (r) between the mean affective ratings of the query images and the corresponding retrieval-based estimates were then computed to evaluate local affective consistency. Although a neighborhood size of five was adopted to provide a stable estimate of the local affective structure, the example-based interface displays only the top three retrieved examples for visualization purposes.

To quantify the retrieval accuracy, an image-wise absolute retrieval error was calculated as the absolute difference between the ground-truth mean affective rating of a query image and its retrieval-based affective estimate. These image-wise errors were used to compute mean absolute error (MAE) values and perform statistical comparisons among the visual representations. Because the distributional assumptions required for the parametric tests were not satisfied, non-parametric statistical tests were employed. Differences between visual representations (ResNet50, ViT-B/16, and CLIP ViT-B/16) were first assessed using Friedman tests for image-wise retrieval errors. When significant main effects were observed, pairwise comparisons were performed using Wilcoxon signed-rank tests with Holm correction for multiple comparisons, and Holm-adjusted p-values below 0.05 were considered statistically significant.

### C. Global Affective Organization

To visualize the global organization of affective representations, feature vectors extracted by CLIP and ViT were embedded using UMAP [57]. Because the CLIP and ViT feature representations have different dimensionalities, UMAP was applied independently to each representation. A two-dimensional embedding with 30 nearest neighbors, cosine distance, and a minimum distance of 0.1 was used for visualization.

An orthogonal Procrustes alignment [58] was subsequently applied to the resulting two-dimensional UMAP embeddings to eliminate arbitrary rotational and reflection differences, thereby facilitating direct visual comparison between the CLIP and ViT representations. The mean valence and arousal ratings were projected onto the aligned embedding spaces for visualization.

For quantitative analysis, spatial autocorrelation was computed directly in the original feature space. Feature vectors were first L2-normalized. For each image, images belonging to the same built-environment group were excluded from the candidate set, and the $k$ nearest remaining images were identified based on cosine distance ($k = 10$). A binary $k$-nearest-neighbor weight matrix was constructed by assigning equal weights to the selected neighbors and was subsequently row-standardized. The same group assignments and exclusion criteria were used for the CLIP and ViT representations. Moran's $I$ and Geary's $C$ were computed using the esda package in the PySAL ecosystem [59]. Higher $I$ values and lower $C$ values indicate stronger affective organization.

To evaluate robustness, repeated subsampling was performed. One hundred subsamples were generated, each containing 1,000 images randomly selected without replacement from the EMOIS dataset; the same subsamples were used for both the CLIP and ViT representations. Moran's $I$ and Geary's $C$ were computed for each subsample, and the mean, standard deviation, and empirical 2.5th–97.5th percentile range across the 100 subsampling runs are reported.

For the cross-dataset comparison, all OASIS images and an equal number of randomly sampled EMOIS images were jointly embedded using UMAP with the same parameter settings as in the within-dataset analysis. For quantitative analysis, 800 images were randomly sampled without replacement from each dataset in each of 100 iterations. The same spatial-autocorrelation procedure and weighting scheme were applied independently to the CLIP representations of both datasets ($k = 10$). For EMOIS, images from the same built-environment group as the query image were excluded from the candidate set before neighbor selection.

### D. Representation–Affect Distance Relationships

To investigate the relationship between representation distance and affective distance, all unique image pairs in the EMOIS dataset were analyzed. Pairs of images belonging to the same environment group were excluded from the primary analysis. For each image pair, cosine similarity between the corresponding image representations was computed according to Eq. (4). The representation distance was then defined as one minus the cosine similarity.

For each image pair, affective distances were computed as the absolute difference in valence ($\Delta V$), absolute difference in arousal ($\Delta A$), and Euclidean distance in the valence–arousal space ($\Delta VA$), defined as

$$\Delta VA = \sqrt{(\Delta V)^2 + (\Delta A)^2}. \tag{5}$$

The relationship between representation distance and affective distance was quantified using Spearman's rank correlation coefficient ($\rho$). Correlations were computed independently for the ViT and CLIP representations, and their difference was defined as

$$\Delta\rho = \rho_{\mathrm{CLIP}} - \rho_{\mathrm{ViT}}. \tag{6}$$

Statistical uncertainty in $\Delta\rho$ was estimated using a group-level jackknife. Each same-environment group was treated as a single jackknife unit, whereas images not belonging to a same-environment group were treated as singleton units. In each jackknife iteration, one unit was omitted along with all image pairs containing any image from that unit. Spearman's rank correlations for the CLIP and ViT representations, as well as their difference ($\Delta\rho$) were recomputed using the remaining image pairs. The variability across the leave-one-unit-out estimates was used to estimate the jackknife standard error of $\Delta\rho$, from which 95% confidence intervals were constructed using the normal approximation.

### E. Example-Based Interpretation Framework

To support human interpretation of predicted emotional impressions, visually similar images were retrieved from the EMOIS dataset. For a given input image, the cosine similarities between its CLIP feature vector [20] and those of all images in the EMOIS dataset were computed. Based on these similarity scores, the three most visually similar images were retrieved together with their mean subjective ratings for valence and arousal obtained from the annotation experiment. The prediction results and retrieved examples were presented using a Flask-based web application. The application displays the predicted valence and arousal values of the input image, together with their associated prediction uncertainties, as bar charts. It also visualizes the subjective ratings of EMOIS images in the VA space and overlays the predicted affective position of the input image to facilitate comparison with human annotations. As a case study, three different spatial images were generated using ChatGPT and used as input examples for the proposed interface.

## V. Results

### A. Affective Characteristics of the EMOIS Dataset

The average number of responses per image for the EMOIS dataset were $120.13 \pm 7.25$ for valence and $120.13 \pm 7.18$ for arousal, excluding the gold-standard images (Table I). For the gold-standard images, the number of responses was 4,205 for both dimensions. The ranges of mean valence and mean arousal scores in the EMOIS dataset were 1.58–6.01, and 2.24–5.85, respectively. For comparison, the OASIS dataset showed ranges of 1.11–6.49 and 1.69–5.72 for valence and arousal, respectively [17]. Despite its restriction to built environments, the EMOIS dataset covers a broad range of emotional ratings that is similar to the range of the OASIS dataset (Fig. 2).

However, the distribution patterns of the two datasets differ substantially. The OASIS dataset exhibited a V-shaped distribution in the VA space, whereas EMOIS images were more densely concentrated in the high-valence and low-arousal region. Specifically, 52.53% of the EMOIS images were located in the lower-right quadrant of the VA space, whereas the remaining images were distributed across the upper-right (17.75%), upper-left (15.93%), and lower-left (13.80%) quadrants. The median standard deviations of valence and arousal ratings in EMOIS were 0.99 and 1.24, respectively, while those of OASIS were 1.09 and 1.68, respectively. This indicates that the variation in arousal ratings among participants was smaller in EMOIS than in OASIS.

To evaluate the consistency of the collected ratings, split-half reliability analysis was performed. This analysis demonstrated high agreement between independently averaged participant groups, with mean split-half reliability coefficients of r = 0.975 (SD = 0.002) for valence and r = 0.937 (SD = 0.004) for arousal, indicating that the aggregated image-level ratings were highly reproducible.

TABLE I
STATISTICS OF THE EMOIS DATASET.

| Item | Valence | Arousal |
|---|---|---|
| Number of images | 1,544 | 1,544 |
| Rating scale | 1–7 | 1–7 |
| Number of valid participants | 4,205 | 4,205 |
| Mean responses / image | $120.13 \pm 7.25$ | $120.13 \pm 7.18$ |
| Median SD across raters | 0.99 | 1.24 |
| Split-half reliability | $0.975 \pm 0.002$ | $0.937 \pm 0.004$ |

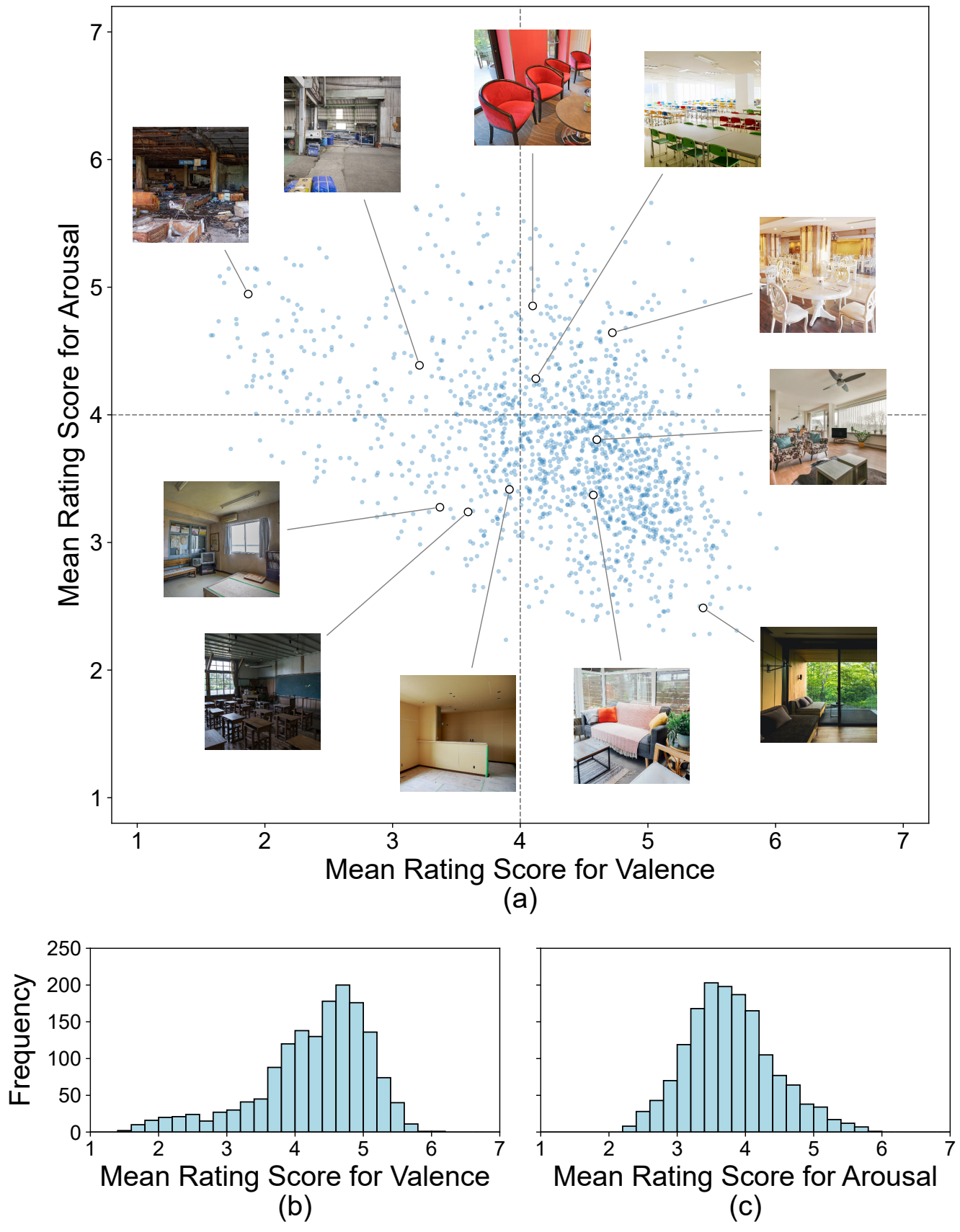

Fig. 2. VA distribution of the EMOIS dataset. (a) Scatter plot of the mean valence and arousal ratings for all images. Ratings represent the mean values obtained from a web-based survey. Dashed lines indicate the midpoint of the rating scale (4, 4). (b) Distribution of valence scores. (c) Distribution of arousal scores.

### B. Spatial Emotion Prediction Performance

Prediction performance was evaluated using repeated group-wise hold-out evaluation. As summarized in Table II, prediction using CLIP representations yielded higher mean $R^2$ values than predictions based on the other evaluated visual representations for both valence and arousal.

The mean $R^2$ values obtained using the CLIP representation were 0.865 (2.5th–97.5th percentile interval: 0.831–0.893) for valence and 0.807 (0.774–0.843) for arousal. The corresponding values for the ViT-B/16 representation were 0.776 (0.731–0.818) and 0.745 (0.689–0.793), whereas those for the ResNet50 representation were 0.707 (0.646–0.766) and 0.679 (0.632–0.720), respectively.

Figure 3 illustrates the relationship between the true and predicted mean ratings for the median repeated hold-out evaluation using the CLIP-based regression model.

Prediction performance was consistently higher for valence than for arousal across all visual representations. As discussed in Section V-A, valence ratings exhibited a broader distribution and lower inter-rater variability than arousal ratings, which may have contributed to the higher prediction performance observed for valence. These results indicated that affective impressions of built environments could be predicted from CLIP representations.

TABLE II
PREDICTION PERFORMANCE OF VISUAL REPRESENTATIONS.

| Backbone | Valence $R^2$ | Arousal $R^2$ |
|---|---|---|
| ResNet50 | 0.707 [0.646–0.766] | 0.679 [0.632–0.720] |
| ViT-B/16 | 0.776 [0.731–0.818] | 0.745 [0.689–0.793] |
| CLIP ViT-B/16 | 0.865 [0.831–0.893] | 0.807 [0.774–0.843] |

Values indicate the mean coefficient of determination ($R^2$). Numbers in brackets indicate the interval between the 2.5th and 97.5th percentiles across 100 repeated group-wise hold-out evaluations.

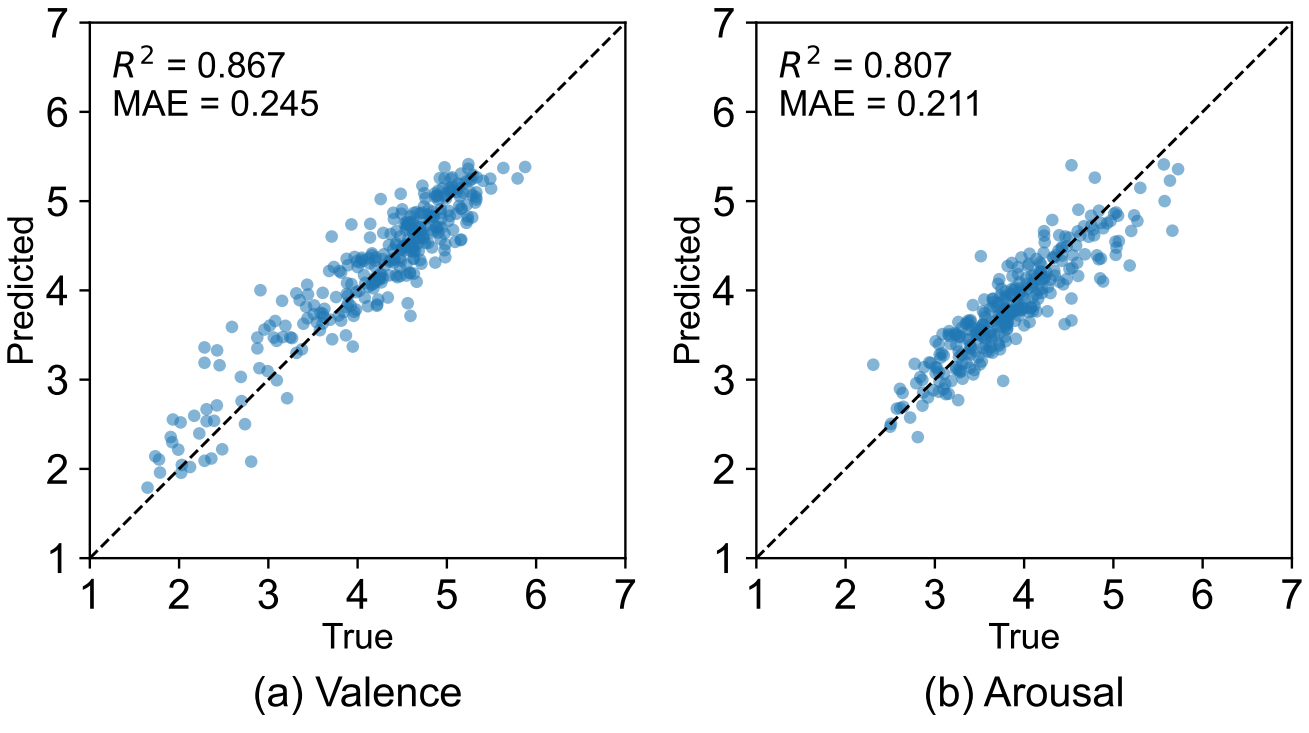

Fig. 3. Comparison of the true and predicted mean ratings for (a) valence and (b) arousal in the median repeated hold-out evaluation of the CLIP-based regression model among 100 repeated hold-out evaluations ($R^2 = 0.867$ and 0.807, respectively).

### C. Local Affective Consistency in Visual Representations

To investigate local affective organization, we compared the retrieval-based affective consistency across ResNet50, ViT, and CLIP representations. For each query image, the top-five nearest neighbors in the representation space were retrieved, excluding images belonging to the same built-environment group as the query image, and their mean affective ratings were compared with the ground-truth ratings of the query image.

Table III summarizes the results. For valence, affective consistency improved from the CNN-based ResNet50 to the transformer-based ViT and further to the multimodal vision–language model CLIP. The correlation coefficients were 0.837, 0.877, and 0.914 for ResNet50, ViT, and CLIP, respectively, whereas the corresponding MAEs were 0.334, 0.299, and 0.260. Statistical comparisons were performed on the image-wise absolute retrieval errors. Friedman test indicated significant differences in retrieval errors among the three representations ($p < 0.001$). Post-hoc Wilcoxon signed-rank tests with Holm correction showed significant differences between all pairs of models (all corrected $p < 0.001$).

For arousal, affective consistency improved from ResNet50 to ViT and CLIP, whereas ViT and CLIP showed comparable performance. The correlation coefficients were 0.767, 0.805, and 0.799 for ResNet50, ViT, and CLIP, respectively, and the corresponding MAEs were 0.314, 0.294, and 0.292. A Friedman test also indicated significant differences in retrieval errors among the three representations ($p = 0.014$). Post-hoc Wilcoxon signed-rank tests with Holm correction indicated that the difference between ViT and CLIP was not significant (corrected $p = 0.622$), whereas both differed significantly from ResNet50 (both corrected $p < 0.01$).

Overall, these findings indicated that transformer-based representations preserved local affective organization more effectively than CNN-based representations. This advantage was particularly pronounced for valence, where CLIP achieved the highest local affective consistency, whereas arousal-related consistency was comparable between ViT and CLIP.

TABLE III
RETRIEVAL-BASED ANALYSIS OF LOCAL AFFECTIVE CONSISTENCY IN VISUAL REPRESENTATIONS.

| | Valence | | Arousal | |
|---|---|---|---|---|
| Backbone | $r$ | MAE | $r$ | MAE |
| ResNet50 | 0.837 | 0.334 | 0.767 | 0.314 |
| ViT-B/16 | 0.877 | 0.299 | 0.805 | 0.294 |
| CLIP ViT-B/16 | 0.914 | 0.260 | 0.799 | 0.292 |

Statistical comparisons were conducted on image-wise retrieval errors using Friedman tests followed by Holm-corrected Wilcoxon signed-rank tests. For valence, all pairwise differences among ResNet50, ViT-B/16, and CLIP ViT-B/16 were significant ($p < 0.001$). For arousal, no significant difference was observed between ViT-B/16 and CLIP ViT-B/16 ($p = 0.622$), whereas both differed significantly from ResNet50 ($p < 0.01$).

### D. Global Affective Organization of Visual Representations

To investigate the global organization of affective information, we first visualized the CLIP and ViT representations using UMAP. The representative UMAP embeddings are shown in Fig. 4. Valence exhibited a relatively smooth apparent gradient in the UMAP visualization, particularly for the CLIP representation. In contrast, arousal exhibited a less continuous pattern, with multiple localized regions rather than a single dominant gradient in both representations.

To quantitatively evaluate these visual observations, spatial autocorrelation was computed using cosine-distance $k$-nearest-neighbor graphs constructed in the original representation space. The corresponding Moran's $I$ and Geary's $C$ statistics are summarized in Table IV. Values are reported as the mean with the corresponding 2.5th–97.5th percentile interval across repeated subsampling runs.

For valence, CLIP achieved substantially higher Moran's $I$ values (0.826 [0.803–0.846]) than ViT(0.736 [0.714–0.763]), indicating stronger spatial organization of valence ratings in the representation space. Likewise, CLIP showed lower Geary's $C$ values (0.209 [0.192–0.225]) than ViT (0.303 [0.284–0.324]), indicating smaller differences in valence ratings between neighboring images.

For arousal, the differences between the two representations were smaller. CLIP exhibited a slightly higher Moran's $I$ value (0.590 [0.563–0.614]) than ViT (0.553 [0.522–0.579]). Geary's $C$ values for CLIP and ViT were similar (0.393 [0.371–0.414] vs. 0.404 [0.381–0.426]).

Overall, the qualitative UMAP visualizations and the quantitative spatial-autocorrelation analysis provided complementary evidence for the fact that the CLIP representation exhibits stronger affective organization, particularly for valence.

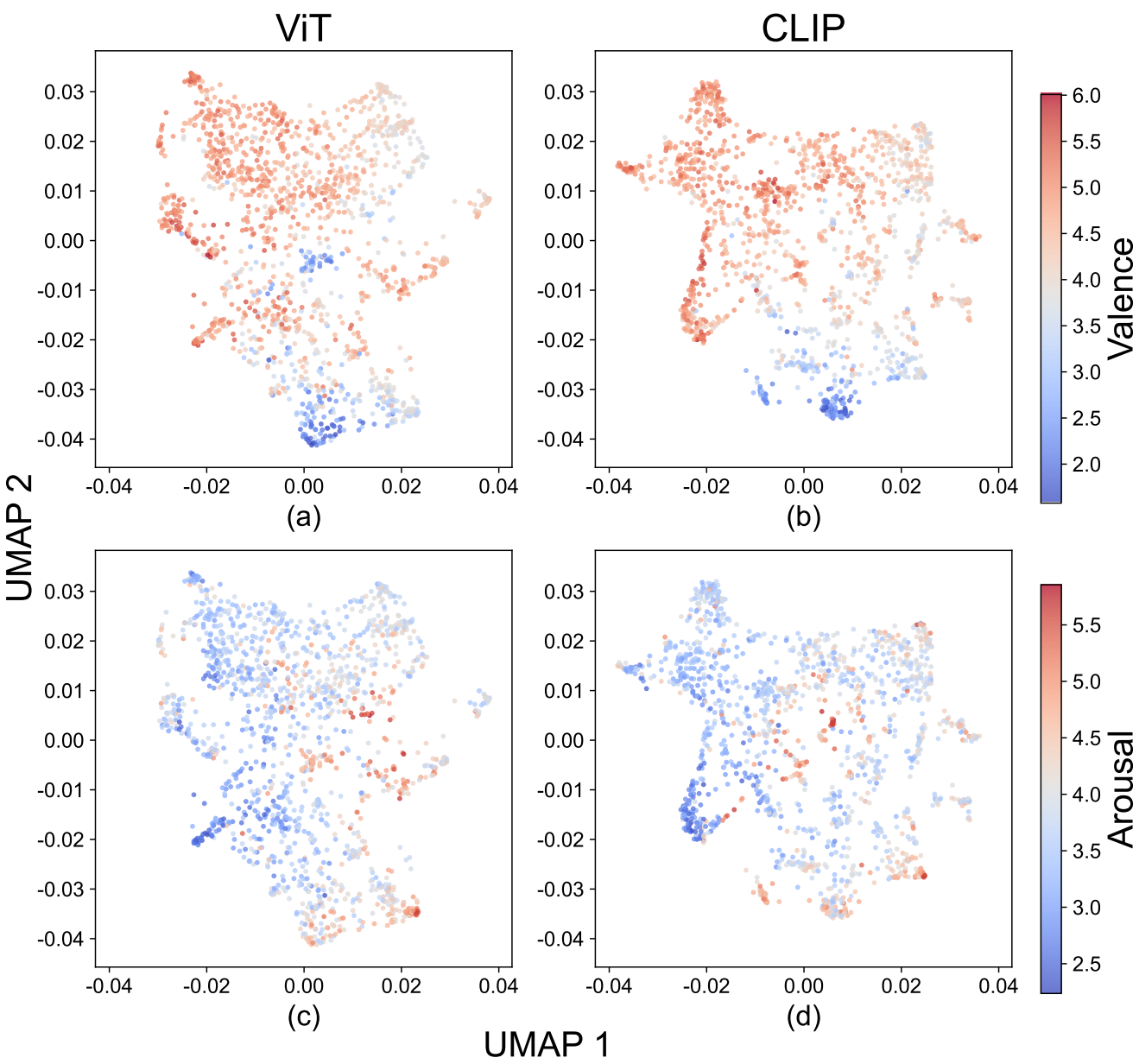

Fig. 4. Representative UMAP embeddings of ViT and CLIP representations for the EMOIS dataset after orthogonal Procrustes alignment. Each point represents an image, colored by its mean valence (top row) or arousal (bottom row) rating. ViT representation results are shown in (a) and (c), and CLIP representation results are shown in (b) and (d).

### E. Relationship Between Representation Distance and Affective Distance

We next examined the relationship between affective distance and representation distance in the ViT and CLIP representation spaces. Fig. 5(a) and (b) correspond to the valence, Fig. 5 (c) and (d) to the arousal, and Fig. 5 (e) and (f) to their combined VA distances, respectively. In both models, affective distance generally increased with increasing representation distance. The strongest relationship was observed for the combined VA distance, followed by valence, whereas the relationship for arousal was weaker. Visually, the CLIP

TABLE IV
QUANTITATIVE ANALYSIS OF GLOBAL AFFECTIVE ORGANIZATION IN CLIP AND VIT REPRESENTATIONS.

| Dimension | Backbone | Moran's $I$ | Geary's $C$ |
|---|---|---|---|
| Valence | ViT | 0.736 [0.714–0.763] | 0.303 [0.284–0.324] |
| | CLIP | 0.826 [0.803–0.846] | 0.209 [0.192–0.225] |
| Arousal | ViT | 0.553 [0.522–0.579] | 0.404 [0.381–0.426] |
| | CLIP | 0.590 [0.563–0.614] | 0.393 [0.371–0.414] |

Values represent the mean across 100 repeated subsampling runs, each using 1,000 images randomly selected without replacement from the EMOIS dataset. The same subsamples were used for both CLIP and ViT representations. Numbers in brackets indicate the empirical 2.5th–97.5th percentile range across the 100 subsampling runs. Spatial autocorrelation was computed using cosine-distance $k$-nearest-neighbor graphs ($k = 10$) constructed in the original feature space, with images belonging to the same built-environment group excluded from the candidate neighbors. Higher Moran's $I$ and lower Geary's $C$ indicate stronger affective organization.

TABLE V
COMPARISON OF SPEARMAN CORRELATION COEFFICIENTS BETWEEN THE EMBEDDING DISTANCE AND AFFECTIVE DIFFERENCE FOR VIT AND CLIP.

| Metric | ViT $\rho$ | CLIP $\rho$ | $\Delta\rho$ |
|---|---|---|---|
| Valence | 0.262 | 0.339 | 0.077 [0.052–0.101] |
| Arousal | 0.195 | 0.201 | 0.006 [-0.017–0.028] |
| VA | 0.330 | 0.408 | 0.077 [0.053–0.101] |

Values represent Spearman's rank correlation coefficients computed using all eligible unique image pairs after excluding pairs of images belonging to the same environment group. $\Delta\rho$ denotes the difference between CLIP and ViT ($\rho_{\mathrm{CLIP}} - \rho_{\mathrm{ViT}}$). Numbers in brackets indicate 95% confidence intervals for $\Delta\rho$ estimated using the group-level jackknife.

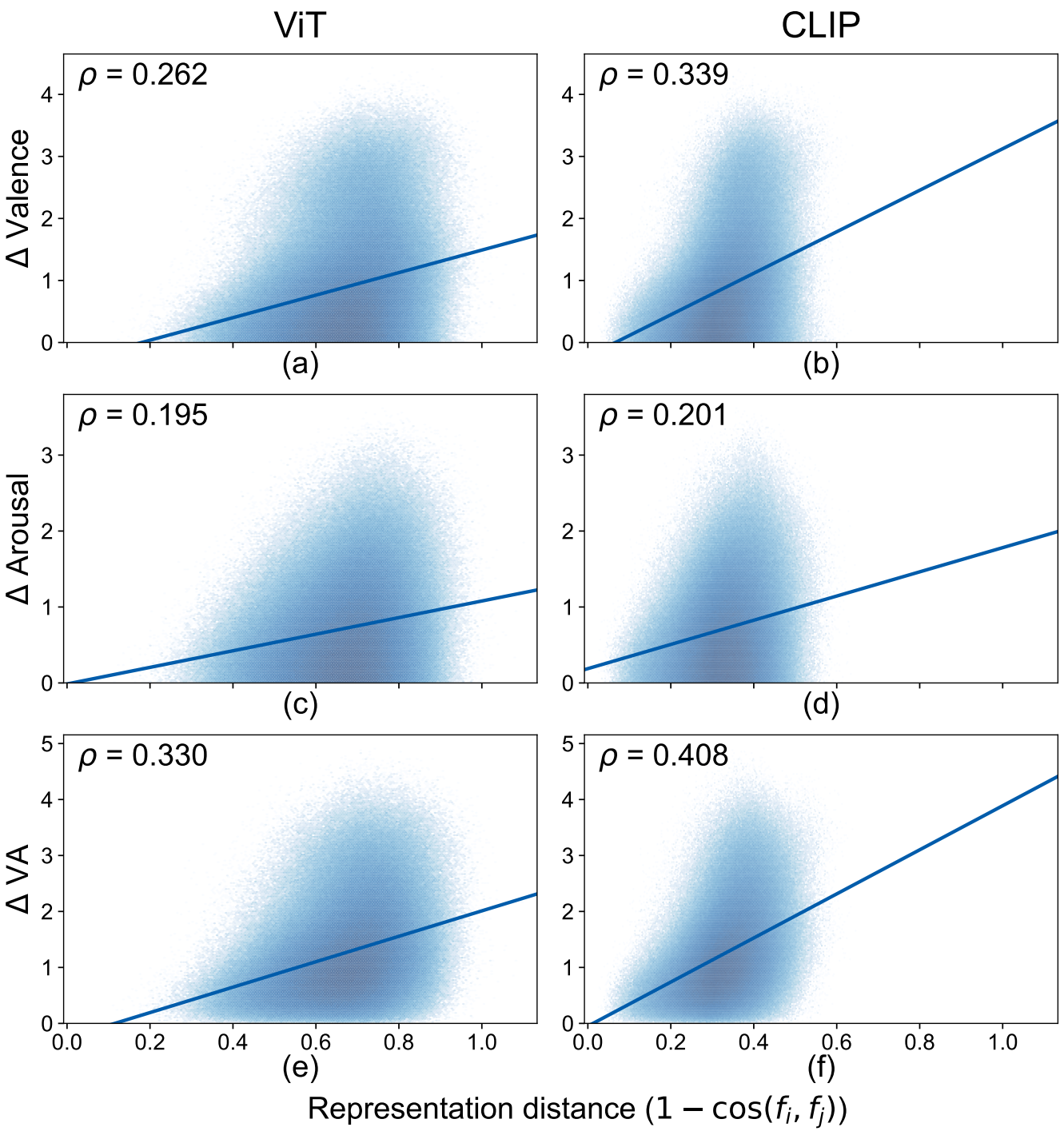

Fig. 5. Relationship between representation distance and affective distance for ViT and CLIP representations. (a) and (b) $\Delta$ valence, (c) and (d) $\Delta$ arousal, and (e) and (f) combined VA distance ($\Delta$ VA). The left and right columns show results obtained using ViT and CLIP representations, respectively. Density plots show the distribution of representation and affective distances for all unique image pairs, excluding within-environment pairs. Solid lines and error bars represent the binned mean affective distance and its standard error of the mean (SEM). Spearman's rank correlation coefficients ($\rho$) computed using the same eligible pair set are shown in each panel.

representations exhibited steeper trends and tighter alignment between representation geometry and affective differences than the ViT representations.

Quantitatively, Table V summarizes the Spearman correlation coefficients computed using all eligible unique image pairs, together with the 95% confidence intervals for the differences between CLIP and ViT estimated using the group-level jackknife. For valence, CLIP showed a stronger association than ViT ($\rho = 0.339$ vs. $0.262$), with an estimated difference of $\Delta\rho = 0.077$ (95% CI: 0.052–0.101). A similar difference was observed for the combined VA distance ($\rho = 0.408$ vs. $0.330$, $\Delta\rho = 0.077$, 95% CI: 0.053–0.101). For arousal, the correlations were similar between CLIP and ViT ($\rho = 0.201$ vs. $0.195$), with a small estimated difference ($\Delta\rho = 0.006$) whose 95% CI included zero (-0.017–0.028).

These findings indicate that CLIP representations exhibited stronger correspondence between representation distance and affective differences than ViT representations for valence and the combined VA distance, whereas no clear difference was observed between the two representations regarding arousal.

### F. Cross-Dataset Comparison of Affective Organization

To compare affective organization between EMOIS and OASIS, we first visualized the CLIP representations of the EMOIS and OASIS datasets using a joint UMAP embedding. The representative embedding is shown in Fig. 6. EMOIS exhibited a pronounced global gradient for valence, consistent with the organization observed in the within-dataset analysis in Fig. 4, whereas the organization of arousal appeared to be less continuous. In contrast, OASIS exhibited relatively smooth global gradients for both valence and arousal, whereas the two datasets occupied partially overlapping regions in the joint embedding.

To quantitatively evaluate these visual observations, spatial autocorrelation was computed independently for each dataset using cosine-distance $k$-nearest-neighbor graphs constructed in the original feature space. As summarized in Table VI, EMOIS exhibited stronger organization of valence ratings than OASIS, with a higher Moran's $I$ (0.815 [0.782–0.841] vs. 0.700 [0.685–0.714]) and a lower Geary's $C$ (0.217 [0.201–0.237] vs. 0.286 [0.274–0.298]).

For arousal, the opposite trend was observed. OASIS exhibited a higher Moran's $I$ (0.740 [0.725–0.753] vs. 0.566 [0.520–0.603]) and a lower Geary's $C$ (0.290 [0.279–0.302] vs. 0.410 [0.381–0.445]) than EMOIS, indicating stronger spatial organization of arousal ratings.

Overall, the UMAP visualization and the quantitative spatial-autocorrelation analysis consistently indicate different patterns of affective organization between the two datasets. EMOIS exhibited substantially stronger organization of valence ratings than arousal ratings, whereas OASIS showed similarly strong organization of both affective dimensions, with slightly stronger organization for arousal ratings.

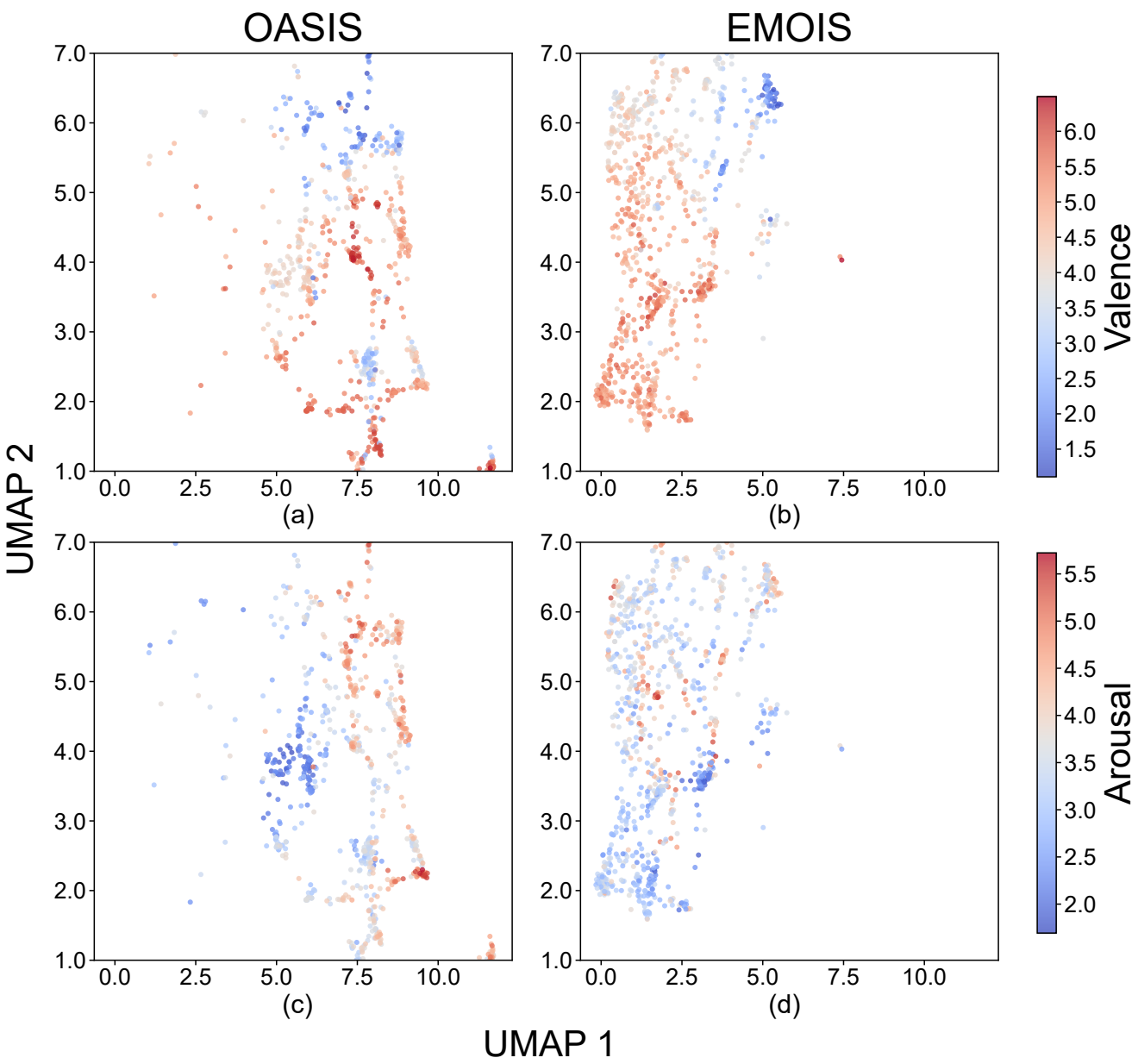

Fig. 6. Comparison of affective organization in the joint UMAP embeddings of CLIP representations between OASIS and EMOIS. (a) and (b) Valence ratings projected onto the joint embedding for OASIS and EMOIS, respectively. (c) and (d) Arousal ratings. Colors indicate mean subjective ratings.

TABLE VI
QUANTITATIVE COMPARISON OF AFFECTIVE ORGANIZATION BETWEEN EMOIS AND OASIS.

| Dimension | Dataset | Moran's $I$ | Geary's $C$ |
|---|---|---|---|
| Valence | OASIS | 0.700 [0.685–0.714] | 0.286 [0.274–0.298] |
| | EMOIS | 0.815 [0.782–0.841] | 0.217 [0.201–0.237] |
| Arousal | OASIS | 0.740 [0.725–0.753] | 0.290 [0.279–0.302] |
| | EMOIS | 0.566 [0.520–0.603] | 0.410 [0.381–0.445] |

Values represent the mean across 100 repeated subsampling runs, each using 800 images randomly selected without replacement from each dataset. Numbers in brackets indicate the 2.5th–97.5th percentile interval. Spatial autocorrelation was computed on cosine-distance $k$-nearest-neighbor graphs ($k = 10$) constructed in the original feature space. For EMOIS, images belonging to the same built-environment group were excluded from the candidate neighbors.

### G. Case Study of the Example-Based interpretation Interface

The preceding analyses showed that the distances in the CLIP representation space are associated with affective differences between spaces in the EMOIS dataset. Based on this observation, we proposed an example-based interpretation interface that retrieves visually similar reference spaces from the EMOIS dataset. Rather than relying solely on numerical predictions, the interface presents the retrieved examples along with their subjective affective ratings, allowing users to interpret the predicted affective characteristics of a query space by comparing them with those of previously evaluated spaces.

Figs. 7–9 show representative case studies of the proposed example-based interpretation interface for different regions of the VA space. Each figure shows a query image (a), predicted valence and arousal values with uncertainty (b), the position of the query image in the EMOIS affective space (c), and the top-three retrieved reference images together with their subjective affective ratings (d)–(f).

In Fig. 7, the query image depicts an office meeting room and was predicted to have moderate valence ($4.05 \pm 0.28$) and moderate-to-low arousal ($3.83 \pm 0.28$). The retrieved reference images consistently exhibited office-like spatial characteristics, including large wooden tables, office-style chairs, open window views, and white interior walls. The cosine similarity values were relatively high (0.93–0.94), and both the query and retrieved images were distributed near the central region of the VA space, indicating visual and affective consistency.

In Fig. 8, the query image depicts a *washitsu* and was positioned in the high-valence and low-arousal region of the EMOIS affective space. The retrieved reference images consistently exhibited Japanese-style spatial characteristics, including low wooden tables, tatami flooring, and wooden structural elements. The retrieved images also exhibited high valence values above 5.0, and low arousal values ranging from 2.42 to 3.05, consistent with the predicted affective tendencies of the query image.

In Fig. 9, the query image depicts an industrial factory-like space characterized by low valence and relatively high arousal. The query is positioned near the edge of the EMOIS affective distribution (Fig. 9(c)), where relatively few reference images exhibited similar affective ratings. The retrieved reference images showed lower cosine similarity values (0.87–0.90) and greater variability in their affective ratings than those in the other case studies. The retrieved reference images also consistently exhibited industrial spatial characteristics.

## VI. DISCUSSION

Our analyses suggest that the examined representations exhibit different affective organization patterns for valence and arousal and between EMOIS and OASIS. Within EMOIS, valence exhibited a more coherent global organization than arousal, whereas comparisons with OASIS suggested additional differences in affective organization between the two datasets.

Consistent with previous findings on CLIP-based aesthetic prediction [12], the CLIP representations outperformed conventional CNN- and ViT-based representations under repeated group-wise hold-out evaluation. Despite the relatively modest size of EMOIS (1,544 images), prediction using CLIP representations achieved mean $R^2$ values of 0.865 for valence and 0.807 for arousal.

One possible explanation for the superior prediction performance of CLIP is the nature of its language-supervised pre-training. Unlike conventional visual representations trained primarily for image classification, CLIP learns the associations between visual patterns and natural language descriptions through image–text alignment. Such pretraining may produce representations that are more closely aligned with human perception and the evaluation of visual content, which may in turn facilitate the modeling of affective responses. This interpretation is supported by previous findings showing that CLIP representations capture subjective perceptual attributes and visually evoked affective responses [12], [13], [15].

Interestingly, differences between CLIP and the comparison representations were more pronounced for valence than for arousal in EMOIS. Across the prediction, retrieval, and representation analyses, valence exhibited more coherent global organization than arousal. One possible interpretation is that perceptual characteristics associated with valence are represented more consistently across built environments, which may have contributed to the stronger predictability and more coherent organization observed for valence. However, the specific perceptual characteristics underlying this tendency remain to be clarified in future work.

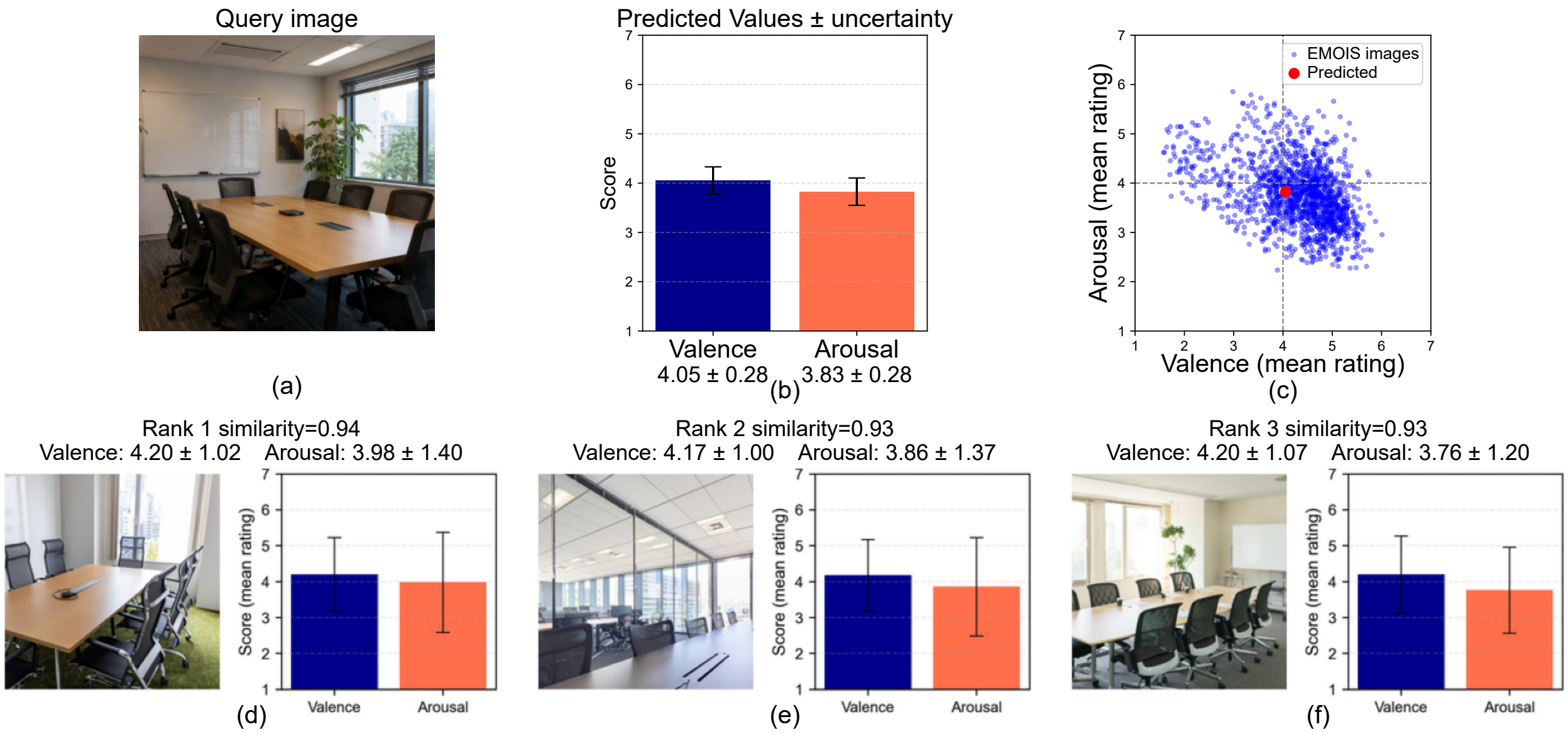

Fig. 7. Case 1: Office meeting room example using the proposed example-based interpretation interface. (a) Query image. (b) Predicted valence and arousal scores with uncertainty estimates. (c) Position of the query image in the EMOIS affective space. The red point indicates the prediction result and blue points indicate EMOIS samples. (d)–(f) Top three most similar reference images retrieved based on CLIP feature similarity, together with their mean valence and arousal ratings (mean ± SD).

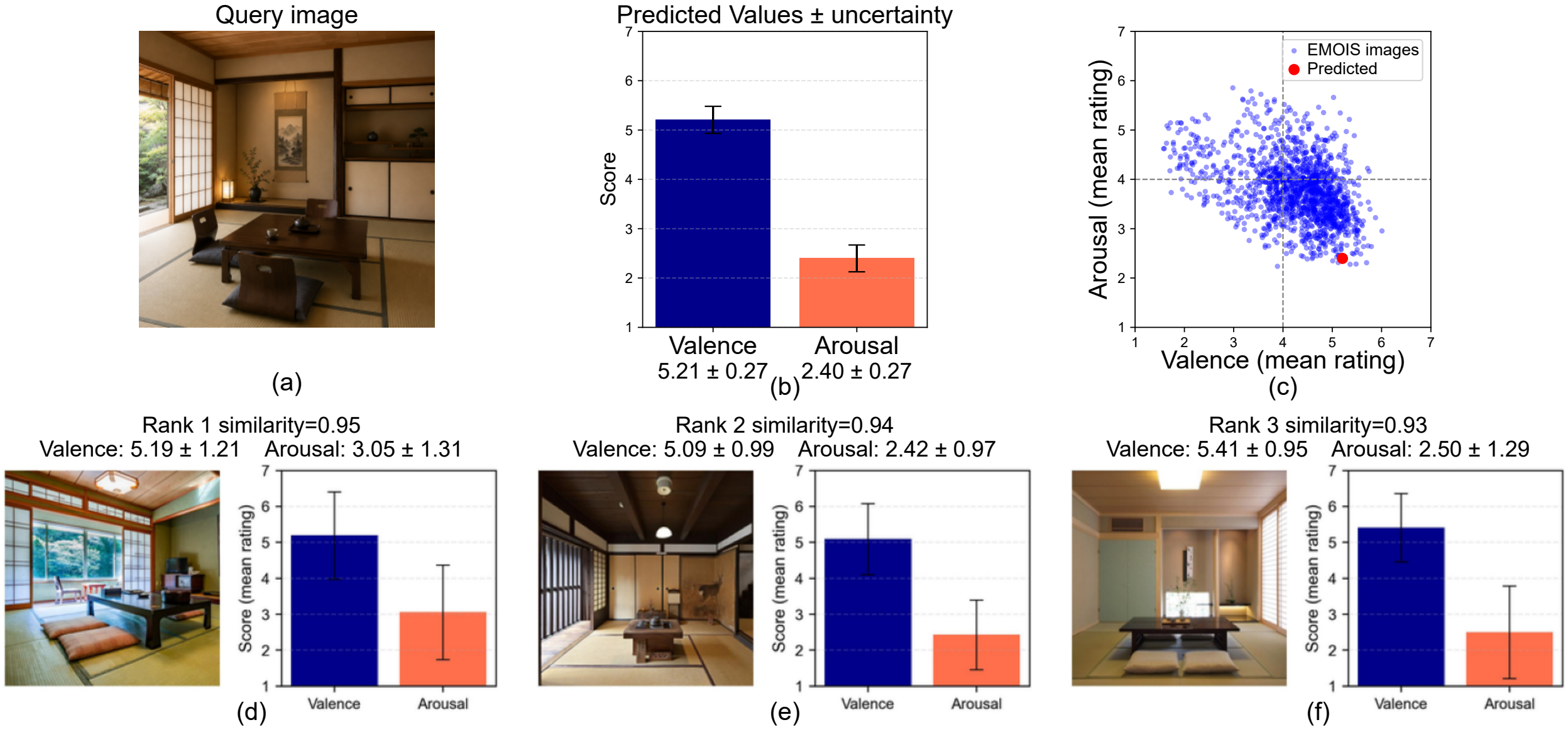

Fig. 8. Case 2: *Washitsu* room example using the proposed example-based interpretation interface. The configuration is the same as in Fig. 7. The retrieved examples consistently reflected calm and low-arousal spatial characteristics.

Comparisons with OASIS further suggested differences in affective organization between the two datasets. While EMOIS showed a stronger global organization for valence than for arousal, OASIS appeared more balanced across the two affective dimensions (Table VI). One possible explanation is that OASIS contains a broader range of highly arousing semantic categories, including violent, sexual, and threat-related scenes, which may have contributed to the stronger organization of arousal observed in that dataset. In contrast, arousal in built environments may depend on combinations of multiple interacting visual characteristics, which may have contributed to the more fragmented global organization observed in EMOIS. However, because EMOIS and OASIS differ in image con-tent, affective distributions, and participant populations, these interpretations remain speculative.

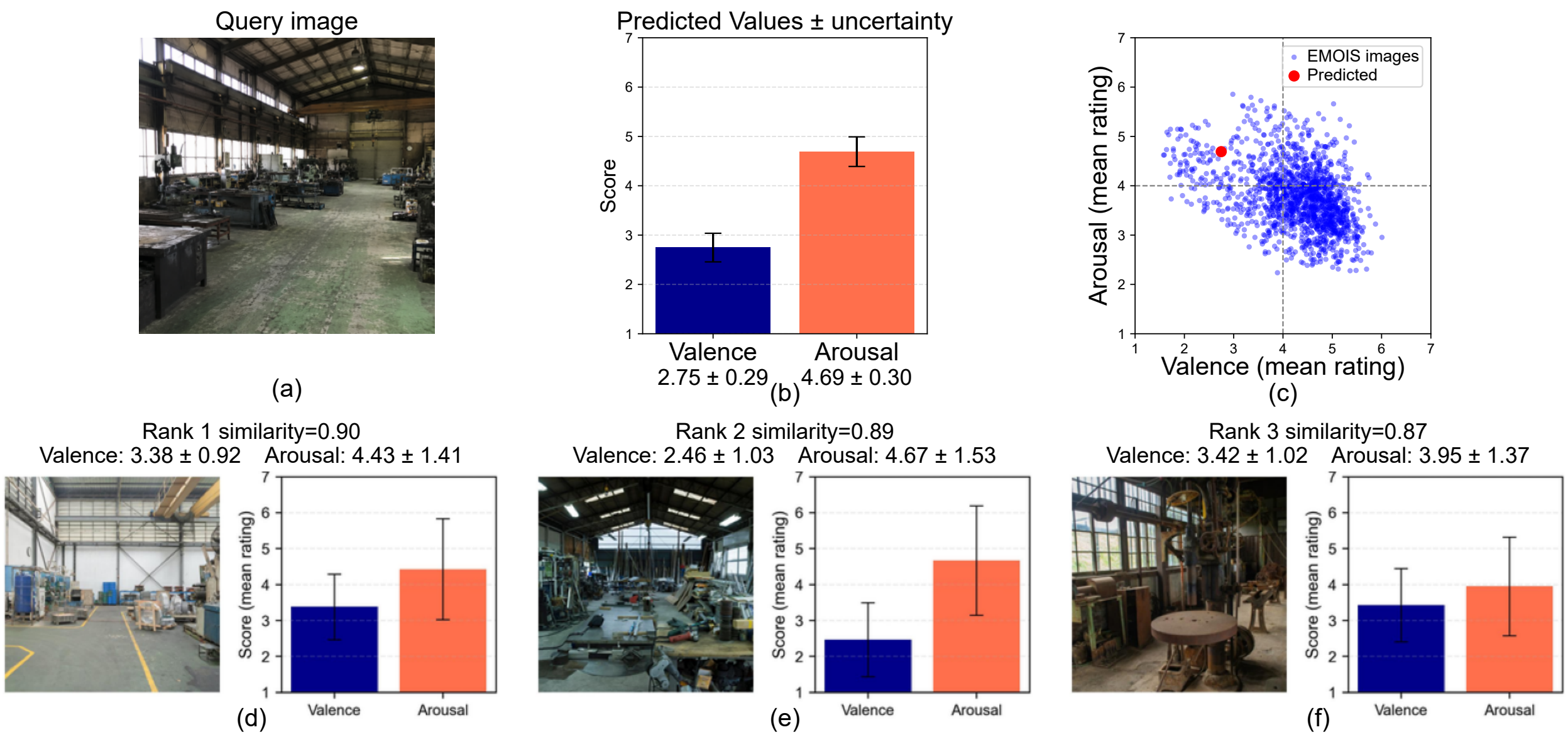

Fig. 9. Case 3: Industrial factory-like space example using the proposed example-based interpretation interface. The configuration is the same as in Fig. 7. The retrieved examples consistently reflected relatively low valence and elevated arousal.

Beyond characterizing affective organization, the retrieval analysis illustrates how providing visually similar reference images along with subjective affective ratings can support interpretation of model predictions. Rather than explaining why a prediction was generated, the proposed approach is intended to help users interpret predicted affective values by presenting visually similar reference spaces with their corresponding subjective affective ratings. Even when visually similar retrieved examples are associated with different affective ratings, users may compare these examples to determine which visual characteristics are associated with differences in the predicted affective ratings. These comparisons may facilitate interpretation of model predictions rather than directly explaining the underlying model. Such example-based presentation may be particularly useful in architectural design, where affective impressions often emerge from the overall spatial atmosphere rather than isolated visual features. However, the practical usefulness of this approach remains to be established through controlled user studies.

Nevertheless, several limitations remain. First, because EMOIS includes culturally specific built environments such as traditional Japanese-style rooms, the annotation experiment was restricted to participants whose nationality and longest residence were both in Japan. Although this design choice reduced variability arising from cultural differences in spatial perception, it may limit the generalizability of the affective ratings to other populations. Second, because EMOIS was constructed using commercially available images, the dataset size remained limited. Larger datasets may further improve prediction performance, particularly for models with higher-dimensional feature representations. Third, although the proposed example-based retrieval approach provides a qualitative means of interpreting predicted affective values, this study did not include a formal user evaluation. Future work should investigate whether retrieved examples improve user trust and the perceived plausibility of affective predictions. Fourth, comparisons between EMOIS and OASIS were based on a single built-environment dataset and a single general affective image dataset. Future studies should examine a broader range of built-environment and general affective image datasets to determine the generality of the observed organizational patterns. Finally, real spatial experiences involve temporal navigation, multimodal perception, and embodied interaction, whereas EMOIS consists only of static visual images. Therefore, extending the framework to include immersive and multimodal spatial affect modeling remains an important direction for future work.

Overall, the present findings suggest that the examined vision–language representations contain structure associated with aggregate affective ratings in EMOIS. Although the present study focused on valence and arousal, emotional experiences may not be fully characterized by low-dimensional affective coordinates alone [27]. Future work could investigate whether foundation-model representations support richer spatial affective experiences, including impressions such as coziness, openness, healing, and tension, and whether these representations can support generation and modification of built environments toward user-specified VA targets. Such systems could provide visual examples of design interventions and help translate affective goals into concrete spatial design suggestions.

## VII. Conclusion

In this study, we introduced EMOIS, a dedicated affective dataset for built environments, and investigated the affective information encoded in CLIP representations. Using CLIP representations, we achieved high prediction performance for affective impressions of built environments under repeated group-wise hold-out evaluation. We further analyzed the organization of affective information within the representation space, revealing distinct organizational patterns for valence and arousal. Comparison with a general affective image dataset further suggested differences in affective organization between the two datasets.

Beyond affective prediction, the proposed example-based retrieval framework illustrates how providing visually similar reference environments together with their subjective affective ratings can support interpretation of model predictions. Future expansion of the dataset could further improve the diversity and reliability of retrieved examples.

Our findings suggest that the examined CLIP representations contain structure associated with aggregate affective ratings. These findings provide a foundation for future studies of affective representation analysis in built environments. Further validation using independent built-environment datasets and cross-cultural annotations will be essential for assessing the robustness and generalizability of the reported findings.

## Generative AI Use Disclosure

OpenAI ChatGPT was used for language editing and grammar refinement during manuscript preparation. They were also used to assist code development and to generate examples of spatial images presented in Section V-G. All experimental procedures, analyses, interpretations, and final manuscript content were reviewed and validated by the authors, who take full responsibility for the work.